\documentclass{article}

\PassOptionsToPackage{numbers, compress}{natbib}

\usepackage[preprint]{neurips_2020}

\usepackage[utf8]{inputenc} 
\usepackage[T1]{fontenc}    
\usepackage{hyperref}       
\usepackage{url}            
\usepackage{booktabs}       
\usepackage{amsfonts}       
\usepackage{nicefrac}       
\usepackage{microtype}      
\usepackage{proofread}

\usepackage{overpic}

\title{Sparse Dynamic Distribution Decomposition: Efficient Integration of Trajectory and Snapshot Time Series Data}

\usepackage{dsfont}
\usepackage{xcolor}
\usepackage{multirow}
\usepackage{booktabs}

\usepackage{amsfonts,amsmath,amssymb,amsfonts}
\usepackage{mathrsfs}

\usepackage[xcolor,framemethod=tikz]{mdframed}
\mdfdefinestyle{tag}{%
    roundcorner=3pt
}

\usepackage{float}

\newcommand{\eps}{\varepsilon}

\newcommand{\tp}[1]{t_{#1}}
\newcommand{\NumCells}[1]{n_{#1}}

\newcommand{\StateSingle}[2]{\vect{x}_{#2,#1}}
\newcommand{\State}{\vect{x}}
\newcommand{\DensityNoArg}{\varrho }

\newcommand{\Density}[2]{\varrho( #1 , #2 ) }
\newcommand{\tDensity}[2]{\tilde{\varrho}( #1 , #2 ) }
\newcommand{\InProd}[1]{\langle #1 \rangle}
\newcommand{\pSum}[1]{S_{#1}}

\newcommand{\ProjPF}{P}

\newcommand{\IndxBasis}{j}

\DeclareMathOperator*{\argmin}{arg\,min}

\newcommand{\vect}[1]{\boldsymbol{#1}}

\newcommand{\Coeff}{c}
\newcommand{\vCoeff}{\vect{\Coeff}}

\newcommand{\p}{\partial}
\newcommand{\dd}{\text{d}}
\newcommand{\de}{\delta}
\newcommand{\De}{\Delta}
\newcommand{\Real}{\mathds{R}}

\newcommand{\MeaSp}{\mathcal{M}}

\newcommand{\PFmat}{P}

\newcommand{\PFc}{\mathcal{P}}

\newcommand{\LimBasis}{N}

\newcommand{\Transpose}[1]{#1^\intercal}

\newcommand{\norm}[1]{\Vert #1 \Vert}

\newcommand{\Basis}{\psi}
\newcommand{\BBasis}{\vect{\psi}}

\graphicspath{{NewFigs/}{../NewFigs/}}

\author{%
Jake P. Taylor-King \\
  Relation Therapeutics\\
  \texttt{jake@relationrx.com} 
  \And
  Cristian Regep \\
  Relation Therapeutics\\
  \texttt{cristian@relationrx.com} 
  \And
  Jyothish Soman \\
  Relation Therapeutics\\
  \texttt{jyothish@relationrx.com} 
  \And
   Flawnson Tong\\
  Relation Therapeutics\\
  \texttt{flawnsontong1@gmail.com} 
    \And
   Catalina Cangea\\
  Relation Therapeutics\\
  \texttt{catalina.cangea@gmail.com} 
    \And
   Charlie Roberts\\
  Relation Therapeutics\\
  \texttt{charlie@relationrx.com}
}

\begin{document}

\maketitle

\begin{abstract}

Dynamic Distribution Decomposition (DDD) was introduced in \citet{taylor2020dynamic} as a variation on Dynamic Mode Decomposition. In brief, by using basis functions over a continuous state space, DDD allows for the fitting of continuous-time Markov chains over these basis functions and as a result continuously maps between distributions. The number of parameters in DDD scales by the square of the number of basis functions; we reformulate the problem and restrict the method to compact basis functions which leads to the inference of sparse matrices only --- hence reducing the number of parameters. Finally, we demonstrate how DDD is suitable to integrate both trajectory time series (paired between subsequent time points) and snapshot time series (unpaired time points). Methods capable of integrating both scenarios are particularly relevant for the analysis of biomedical data, whereby studies observe population at fixed time points (snapshots) and individual patient journeys with repeated follow ups (trajectories). 

\vspace{0.5em}
Code to be made available at: \url{https://github.com/RelationRx/SparseDDD}

\end{abstract}

\section{Introduction}

Consider two scenarios, the first whereby a set of (stochastic) sample path trajectories are observed
\begin{equation}\label{eq_set_Traj}
    \mathcal{T} = \left\{ \vect{X}_k ( t_r ) \in \mathds{R}^d : \vect{X}_k ( t_{r} ) = \vect{X}_k ( t_{r-1} )  + \int_{t_{r-1}}^{t_{r}} \vect{f}( \vect{X}_k (s) ) \dd Y(s) \,\text{ for }\, \begin{array}{c}
         r = 1,\dots,R_\mathcal{T} \\
         k =1,\dots,K_\mathcal{T}
    \end{array} \, \right\} \, ,
\end{equation}
and the second where snapshots are reported at fixed points in time from the same initial state
\begin{equation}\label{eq_set_Snap}
    \mathcal{S} = \left\{ \vect{X}_{k_r} ( t_r ) \in \mathds{R}^d : \vect{X}_{k_r} ( t_{r} ) = \vect{X}_{k_r} ( t_0 )  + \int_{t_0}^{t_{r}} \vect{f}( \vect{X}_{k_r} (s) ) \dd Y(s) \,\text{ for }\, \begin{array}{c}
         r = 1,\dots,R_\mathcal{S} \\
         k_r \in\mathds{N}_+
    \end{array} \, \right\}\, .
\end{equation}
In equations \eqref{eq_set_Traj}--\eqref{eq_set_Snap}, the integrals are over a suitable class of random process, e.g., a pathwise Stieltjes integral for H\"{o}lder continuous random processes $\vect{X}(t)$ and $Y(t)$ \cite{chen2019pathwise}. For simplicity, a stochastic differential equation (SDE) model would suffice --- although the methodology presented applies to more general stochastic processes. We present a method to integrate both $\mathcal{T}$ and $\mathcal{S}$ for a predictive time series model.

The ability to integrate different sources and formats of temporal data is hugely important to the biomedical community. For a topical example, early data from patients with COVID-19 infections has led to the single cell characterisation \cite{sungnak2020sars} of various tissues at different stages (and outcomes) during infection, such as healthy (susceptible), early infection, the pulmonary phase, the hyper-inflammatory phase and finally leading to: a healthy (recovered) state, a fibrotic state, or even death \cite{siddiqi2020covid}. However these are \emph{not} the same patients being followed longitudinally --- only recently has data started to become available for the monitoring of individual patient journeys \cite{zhao2020longitudinal}. While there may have been differences in the collection, preprocessing and quality control protocols that need addressing, hypothetically said datasets should be combined such that the resulting analysis leads to higher quality inferences than any one dataset alone.

In experimental and clinical settings, discrete-time models, such as Dynamic Mode Decomposition (DMD) \cite{tu2013dynamic, williams2015data, schmid2010dynamic}, are unsuitable for the unequally spaced observations that typically occur in real world experimental scenarios \cite{de2017discrete}. Countless methods for time series analysis have been developed for trajectory data, often based on mechanical or phenomenological ODE \cite{ aldridge2006physicochemical, yang2019complex, rubanova2019latent} or SDE \cite{mariani2016stochastic, duan2009modeling} models in various scenarios, e.g., patient data \cite{beaulieu2018mapping} or signalling networks \citep{zechner2012moment}. Methods for snapshot time series are a more recent phenomenon due to the wide availability of \emph{``omic''} data that typically requires a destructive measurement process \cite{zunder2015continuous}. These methods involve mapping distributions between time points, for example via minimising earth mover's distance between subsequent time points \cite{Schiebinger_2017}, or inferring a potential well \cite{hashimoto2016learning}. To clarify terminology, \emph{pseudotime methods} for inferring trajectories of differentiating cells within a heterogeneous population for a single snapshot exist \cite{setty2016wishbone, ching2018opportunities, Saelens276907,cannoodt2016computational}. Finally, in the work by \citet{weinreb2018fundamental}, it is shown there are limits on what may be inferred from snapshot data alone --- there are too many models that are hypothetically possible. Therefore, we posit that approaches integrating both trajectory and snapshot data (when available) should be utilised.

In this work, we propose a sparsity-promoting modification on Dynamic Distribution Decomposition (DDD) \cite{taylor2020dynamic} --- itself an extension to DMD. DMD and related approaches are emerging powerful methods for (often discrete) time series analysis capable of various tasks: analysis of key behaviours \cite{schmid2011applications}, anomaly/error detection \cite{taylor2020dynamic}, control \cite{klus2020data}, parameter estimation \cite{mauroy2017koopman, riseth2017operator} and scaling into high dimensions \cite{klus2018tensor}. DDD is essentially the numerical approximation of the Perron--Frobenious operator \cite{klus2017data,klus2016On} in continuous time, with constraints to enforce positivity and conservation of mass. In contrast, DMD approximates the Koopman operator. In more general terms, DDD fits a continuous-time Markov chain over a set of basis functions, thus learning an autonomous dynamical system. This approach is highly general and does not distinguish between trajectory and snapshot datasets. However, the Markov rate matrix is dense and therefore computationally expensive to infer for a large number of basis functions. 
We show that by limiting the behaviour of the Perron--Frobenius operator to local differential operators only, we can recast the problem such that one only has to  infer a sparse matrix. 

In Section \ref{sec_method}, we present the method for both DDD and Sparse DDD. In Section \ref{sec_workedexample}, we apply the method to a synthetic dataset generated by an SDE and show that we can integrate a small snapshot dataset with a small trajectory dataset and obtain reduced errors below either dataset alone. Finally, we provide a discussion of the methodological advantages and disadvantages in Section \ref{sec_discussion}.

\section{Method}\label{sec_method}

One of the most general mathematical set ups for modelling the evolution of a probability density involves the specification of a hyperbolic (transport) or parabolic partial differential equation and then solving forward in time. 
The most general of such equations involves the Perron--Frobenius operator $\PFc$ (in continuous time)
\begin{align}\label{eq_continuous_PF}
  \frac{\p }{\p t} \Density{ t }{ \State }  = \PFc \Density{ t }{ \State }  \, ,
\end{align}
with constraints $\int_{\MeaSp} \Density{ t }{ \State }  \dd\State = 1$ and $\Density{t}{\State}\geq 0 \, $ and initial condition $\Density{ t = 0 }{ \State } = \varrho_0(\State)$. Said operators can account for many dynamical systems --- see examples.

\begin{figure}[h]
\begin{mdframed}[style = tag]
\begin{center}
    \subsection*{Example 1: Stochastic Matrices as Perron--Frobenius Operators}
\end{center}
Consider a discrete-time Markov chain over a countable number of discrete states $i\in\mathcal{I}\subset \mathds{Z}$ with recursive relation
  \begin{align}\label{eq_simple_exp1}
  c_i^{(k+1)} = \sum_{j\in\mathcal{I} } P_{ij} c_j^{(k)} \, , \qquad k = 0,1,2,\dots\, ,
  \end{align} 
  then $P_{ij}$ is a left stochastic matrix with $P_{ij} = \Pr ( X_{k+1} = i \vert X_{k} = j )$ and $\sum_{i\in\mathcal{I}} P_{ij} = 1$. In this case, $P$ is the Perron--Frobenius operator. 
  
  
\begin{center}
    \subsection*{Example 2: Fokker--Planck Equations as Perron--Frobenius Operators}
\end{center}
As an alternative, consider continuous-time stochastic differential equation $\{ \vect{X}_t \in \mathds{R}^d : t \geq 0\}$
\begin{equation}
 \dd \vect{X}_t = \vect{\mu}(\vect{X}_t,t)\, \dd t + \vect{\sigma} ( \vect{X}_t,t )\, \dd\vect{W}_t,
\end{equation} 
for $\vect{\mu} \in \mathds{R}^d$ and $ \vect{\sigma}(\vect{X}_{t},t) \in \mathds{R}^{d \times d'}$ where $\vect{W}_{t}$ is a $d'$-dimensional standard Wiener process. By defining diffusion tensor $\mathbf{D}=\frac{1}{2}\vect{\sigma}\Transpose{\vect{\sigma}}$, we write the Fokker--Planck equation, a partial differential equation for the probability density $\varrho(t,\State)$ that $\vect{X}_t \in [\State, \State + \De \State)$ as
 \begin{equation}\label{eq_FP}
 \frac{\partial \varrho(\vect{x},t)}{\partial t} = -\sum_{i=1}^{d} \frac{\partial}{\partial x_i} \left[ \mu_i(\vect{x},t) \varrho(\vect{x},t) \right] + \sum_{i=1}^{d} \sum_{j=1}^{d} \frac{\partial^2}{\partial x_i \, \partial x_j} \left[ D_{ij}(\vect{x},t) \varrho(\vect{x},t) \right] \, .
 \end{equation}  
By comparing equation \eqref{eq_continuous_PF} to equation \eqref{eq_FP}, we see that the right hand side to the Fokker--Planck equation is the Perron--Frobenius operator.
\paragraph{Link to transition kernel}
Finally, we can build intuition for operator $\PFc$ by linking kernel function $ k_{t} (\State, \vect{y}) = \Pr ( \vect{X}_t  = \State | \vect{X}_0 = \vect{y})$ to the exponentiated operator via the equation
\begin{align}
    e^{t\PFc} \varrho_0(\State)
    =\int_{\MeaSp} k_{t}(\State,\vect{y}) \varrho_0(\vect{y}) \dd \vect{y} \, .
\end{align} 
\end{mdframed}
\vspace{-1em}
\end{figure}

\subsection{Finite Dimensional Approximation}\label{sec_fin_dim_appr}
We need to find a finite dimensional approximation of $\PFc$; we achieve this via non-negative basis functions $\BBasis(\State) =  \Transpose{[ \Basis_{1}(\State) , \dots , \Basis_{\LimBasis}(\State) ]}$ that we specify as probability density functions; whereby $\int_{\MeaSp} \Basis_{\IndxBasis} (  \State ) \dd \State=1$. For all $t \in (\tp{1},\tp{R})$ we take the ansatz that we can expand $\DensityNoArg$ as a Galerkin approximation
\begin{align}\label{eq_form_Density}
  \tDensity{ t }{ \State } = \Transpose{\vCoeff}(t) \BBasis ( \State ) = \sum_{\IndxBasis = 1}^{\LimBasis} \Coeff_{\IndxBasis}( t ) \Basis_{\IndxBasis} (  \State ) \, ,
\end{align}
where $\vCoeff(t) = \Transpose{[ \Coeff_{1}(t) , \dots , \Coeff_{\LimBasis}(t) ]}$. For conservation of mass we require $\sum_{i=1}^N c_i(t) = 1$. A tilde ($\sim$) over $\DensityNoArg$ denotes when this approximation is used.

We then derive a linear system of ODEs for the coefficients $\vCoeff(t)$ by noting in weak form \citep{Ziemer_2012,Gilbarg_2015} equation \eqref{eq_continuous_PF} is
\begin{align}
  \InProd{ \Basis_i , \dot{\DensityNoArg} }_{\mu} = \InProd{ \Basis_i , \PFc\DensityNoArg }_{\mu} \quad\text{for}\quad \InProd{f,g}_{\mu} = \int_{\MeaSp} f(\State) g (\State) \,  \dd \mu( \State )\, ,
\end{align}
and therefore
\begin{align}\label{eq_MQ_PF_form}
  M \dot{\vCoeff}(t) = Q  \vCoeff (t) \quad \text{for}\quad \begin{array}{l} M_{ij} = \InProd{\Basis_i, \Basis_j}_{\mu} \\  Q_{ij} = \InProd{\Basis_i, \PFc\Basis_j}_{\mu}. \end{array}
\end{align}
\begin{figure}[h]
\begin{mdframed}[style = tag]
\begin{center}
\subsection*{Calculation of Galerkin Approximation Coefficients for PDFs}
\end{center}
When basis functions are probability density functions, we can find the values of $\vCoeff_r$ at each time point by noting that the value of $\Coeff_{\IndxBasis} ( \tp{r} )$ must be proportional to the probability that basis function $\IndxBasis$ created the data at that time point, therefore
\begin{align}
  \Coeff_{\IndxBasis} ( \tp{r} ) =  \sum_{i = 1}^{\NumCells{r}} \Basis_{\IndxBasis} ( \StateSingle{i}{r} ) \bigg/ \sum_{\IndxBasis = 1}^{\LimBasis} \sum_{i = 1}^{\NumCells{r}} \Basis_{\IndxBasis} ( \StateSingle{i}{r} ) \, .
\end{align}
\end{mdframed}
\vspace{-1em}
\end{figure}

\subsection{Construction of Loss Function} \label{sec_PF_matrix_sel}
We now motivate a choice in loss function analogous to least squares fitting error for ODEs. Assuming that $M$ is invertable, then we define the projection of $\PFc$ onto the basis functions as
\begin{align}
  \ProjPF := M^{-1} Q \, .
\end{align}
We can solve the system of ODES in equation \eqref{eq_MQ_PF_form} via the matrix exponential operation, specifically
\begin{equation}\label{eq_ODE_sol}
  \vCoeff  ( t ) = e^{t \ProjPF }   \vCoeff_*  \, ,
\end{equation}
where $\vCoeff_*$ are the coefficients at corresponding to the initial condition $\tilde{\varrho}_0 = \Transpose{\vCoeff_{*} } \BBasis ( \State )$. In equation \eqref{eq_ODE_sol}, we allow mis-specification of the initial condition $\varrho_0 = \varrho_0(\State)$ by specifying that $\vCoeff_*$ are free parameters within model. We can use the $L^2$ norm to give a measure of how well $\PFc$ represents the evolution of the densities. The squared relative prediction error at time $\tp{} = \tp{r}$ for $r = 0,\dots, R$ is
\begin{align}\label{eq_inf_dim_error}
\epsilon_r^2	( \PFc , \varrho_0 ) &:=  \frac{ \norm{ e^{\tp{r}\PFc }  \varrho_0 ( \cdot ) -  \Density{\tp{r}}{\cdot} }_{L^2}^2 }{ \norm{  \Density{\tp{r}}{\cdot}  }_{L^2}^2 } = \frac{ \int_{\MeaSp} \left[ e^{ \tp{r} \PFc} \varrho_0 ( \State )  -  \Density{\tp{r}}{\State} \right]^{2} \dd\State  }{  \int_{\MeaSp} \left[  \Density{\tp{r}}{\State} \right]^{2} \dd\State  } \, .
\end{align}
We then define the mean squared relative prediction error as $ \sum_{r=0}^{R} \epsilon_r^2 ( \PFc , \varrho_0 ) /(R+1)$. Were this computationally possible, we would like to infer the differential operator and initial state
  \begin{align}\label{eq_ideal_scen}
  \{   \PFc , \varrho_0  \} :=
           \argmin_{\PFc , \,  \varrho_0  \geq 0, \varrho_0\in L^1 } \frac{1}{R+1}
          \sum_{r=0}^R \epsilon_r^2 (   \PFc , \varrho_0  ) \, .
          \end{align}
However, we must use the finite dimensional Galerkin approximation; therefore we approximate
\begin{align}
 \epsilon_r^2 (   \PFc , \varrho_0   )  \stackrel{\text{Eq. (\ref{eq_form_Density} ) } }{\approx} \eps_r^2	( \ProjPF ,  \vCoeff_{*}   ) \, .
\end{align}
and we calculate the time $\tp{} = \tp{r}$ error as 
\begin{align}
\eps_r^2	( \ProjPF ,  \vCoeff_{*}   ) := \frac{ \norm{ \Transpose{(e^{\tp{r}\ProjPF} \vCoeff_{*}  - \vCoeff_{r}  ) } \BBasis ( \cdot ) }_{L^2}^2 }{ \norm{  \Transpose{\vCoeff_{r} } \BBasis ( \State )  }_{L^2}^2 }  = \frac{ \norm{ e^{\tp{r}\ProjPF} \vCoeff_{*}  - \vCoeff_{r}   }_{M}^2 }{ \norm{ \vCoeff_{r}  }_{M}^2 }  = \frac{ \Transpose{ [ e^{ \tp{r} \ProjPF} \vCoeff_{*} -  \vCoeff_{r}]} M [ e^{ \tp{r} \ProjPF} \vCoeff_{*} -  \vCoeff_{r}]    }{ \Transpose{ \vCoeff_{r} } M  \vCoeff_{r}   } \, ,
\end{align}
where we introduce the mass matrix weighted norm $\norm{\vect{c}}_M := \sqrt{ \Transpose{\vect{c}}M \vect{c}}$. We then choose the loss function
\begin{equation}
\mathcal{L} =  \frac{1}{R+1}  \sum_{r=0}^R    \eps_r^2 (  \ProjPF , \vCoeff_{*} )  \, .
\end{equation}
\subsubsection{Option 1: Inference of Projected Perron--Frobenius Operator Matrix $P$ [DDD]}\label{sec_DDD}
We now need to address the issue of how to determine $\ProjPF$ from data. In order to preserve probability density and positivity, we require $\ProjPF \in \mathscr{P}$ for
\begin{align}\label{eq_req_mass_conservation}
\mathscr{P} = \left\{   \ProjPF \in \Real^{\LimBasis\times\LimBasis} : \sum_{k=1}^{\LimBasis}  \ProjPF_{kj} = 0 \,\text{ and } \,   \ProjPF_{ij} \geq 0 \text{ for } i,j = 1,\dots,\LimBasis \text{ and } i\neq j \right\} \, .
\end{align}
This is motivated by conservation of probability mass, i.e., requiring $\dd / \dd t [\norm{ \vect{c} (t)}_1] = 0$ and employing equation \eqref{eq_MQ_PF_form}. DDD then corresponds to finding the solution to a finite dimensional approximation to equation \eqref{eq_ideal_scen}, given by
\begin{align}\label{eq_obj_func}
\{  \ProjPF ,  \vCoeff_{*} \} :=
\argmin_{\ProjPF\in\mathscr{P},\, \vCoeff_{*} \in\Lambda } \frac{1}{R+1}  \sum_{r=0}^R    \eps_r^2 (  \ProjPF , \vCoeff_{*} )  \, .
\end{align}
From \citet{taylor2020dynamic}, the matrix $P$ and initial condition $ \vCoeff_{*}$ with relevant constraints have $(\LimBasis - 1) (\LimBasis + 1) = \mathcal{O}(\LimBasis^2)$ degrees of freedom. To enable gradient descent, one can calculate the $\tp{r}$ relative error with respect to $\ProjPF$ as
\begin{align}\label{eq:err_grad_pf}
\frac{\p \eps_r^2 }{ \p \ProjPF }  &=\frac{2}{  \Transpose{ \vCoeff_{r} } M  \vCoeff_{r}   } \sum_{k=0}^\infty \frac{ t^{k+1} }{(k+1)!} \sum_{j=0}^k \Transpose{{(\ProjPF^{k-j})}}M(e^{\tp{r} \ProjPF } \vCoeff_{*} -  \vCoeff_{r} ) \Transpose{  \vCoeff_{*} }\Transpose{{(\ProjPF^j)}}	\\ 
&=  \frac{2}{  \Transpose{ \vCoeff_{r} } M  \vCoeff_{r}   }  \sum_{k=0}^\infty \frac{ t^{k+1}}{(k+1)!} \pSum{k} \, ,
\end{align}
and the derivative with respect to $\vCoeff_{*}$ as
\begin{align}
\frac{\p \eps_r^2	}{ \p  \vCoeff_{*}  } =  \frac{2}{ \Transpose{ \vCoeff_{r} } M \vCoeff_{r}   } \Transpose{ [ e^{\tp{r} \ProjPF} ] } M [ e^{\tp{r} \ProjPF} \vCoeff_{*} -  \vCoeff_{r}] \, .
\end{align}
The terms $\pSum{k} =  \sum_{j=0}^k \Transpose{{(\ProjPF^{k-j})}}M(e^{\tp{r} \ProjPF} \vCoeff_{*} -  \vCoeff_{r} ) \Transpose{ \vCoeff_{*} }\Transpose{{(\ProjPF^j)}} \,$
can be calculated using the recursion relation
\begin{align}\label{eq_recusion_rel}
\pSum{k} =   \Transpose{\ProjPF} \pSum{k-1} + \pSum{k-1}
  \Transpose{\ProjPF} - \Transpose{\ProjPF} \pSum{k-2}
  \Transpose{\ProjPF}\, , \quad \text{where} \quad 
 \left\{\begin{array}{l} \pSum{-1} = {\bf 0} \\  \pSum{0} = M(e^{\tp{r} \ProjPF }
  \vCoeff_{*} -  \vCoeff_{r}) \Transpose{ \vCoeff_{*} }  \, . \end{array} \right.
\end{align}



\subsubsection{Option 2: Inference of Matrix $Q$ for Local Differential Operators [Sparse DDD]}\label{sec_SDDD}
Another option is to use compact basis functions (i.e., $\Basis_i$ has finite support), then by restricting $\PFc$ to local differential operators only, $M$ can be made sparse. For the sake of argument, $M$ could have $\mathcal{O}(\LimBasis \log \LimBasis)$ non-zero elements; however $\ProjPF = M^{-1}Q $ would still have $\mathcal{O}(\LimBasis^2)$ degrees of freedom (because $Q$ is dense). A workaround would be to enforce that $\PFc$ consists only of \emph{local} differential operators, namely
\begin{equation}\label{eq_PF_local}
    \PFc =  \sum_{ |\vect{\alpha}| \in \mathds{N}} \sum_{\vect{\alpha} \in \mathds{N}_0^d} a_{\vect{\alpha} } \frac{\p^{ | \vect{\alpha} | }}{ \p x_1^{\alpha_1} \dots \p x_d^{\alpha_d} } \, ,
\end{equation}
for $a_{\vect{\alpha} } \in \mathds{R}$. For the case of equation \eqref{eq_PF_local}, whenever the mass matrix is zero, the corresponding entry of $Q$ is also zero: 
\begin{equation}
    M_{ij} = 0 \quad \implies \quad Q_{ij} = 0 \, .
\end{equation}
Therefore, DDD can be made sparse by only inferring the non-zero entries of matrix $Q$, thus Sparse DDD. We then simply change the argument of the minimisation problem, so 
\begin{align}\label{eq_obj_func_Q}
\{  Q,  \vCoeff_{*} \} := \argmin_{Q\in\mathscr{Q},\, \vCoeff_{*} \in\Lambda } \frac{1}{R+1}  \sum_{r=0}^R    \eps_r^2 (  Q , \vCoeff_{*} )  \, .
\end{align}
and therefore we can calculate the derivative with respect to $Q$ using the chain rule
\begin{equation}
    \frac{\p \eps_r^2 }{ \p Q}  = M^{-1}   \frac{\p \eps_r^2 }{ \p \ProjPF } \, .
\end{equation}
We then also have to modify equation \eqref{eq_req_mass_conservation} to choosing \begin{align}\label{eq_req_mass_conservation_Q}
\mathscr{Q} = \left\{   Q \in \Real^{\LimBasis\times\LimBasis} : \sum_{k=1}^{\LimBasis} \sum_{l=1}^{\LimBasis} M^{-1}_{kl} Q_{lj} = 0 \,\text{ and } \, \sum_{l=1}^{\LimBasis}  M^{-1}_{il} Q_{lj} \geq 0 \text{ for} \begin{array}{c}
   i,j = 1,\dots,\LimBasis   \\
   i\neq j
\end{array}\right\} \, .
\end{align}

\subsection{Integrating Trajectory and Snapshot Data}\label{sec_integration}
It is simple to calculate a series of coefficients for both the collection of $K_\mathcal{T}$ trajectory time series, $ \{ \vect{c}_{r,k}\}_{r=1, k = 1}^{R_\mathcal{T},K_\mathcal{T}}$, and the snapshot dataset $ \{ \vect{c}_r \}_{r=1}^{R_\mathcal{S}}$. We combine loss functions to obtain
\begin{equation}
      \frac{\lambda}{ K_\mathcal{T} ( R_\mathcal{T}+1 )}  \sum_{k=0}^{K_\mathcal{T}} \sum_{r=0}^{R_\mathcal{T}}    \eps_{r,\mathcal{T}}^2 (  \ProjPF , \vCoeff_{k,*} ) +  \frac{1-\lambda}{R_\mathcal{S}+1}  \sum_{r=0}^{R_\mathcal{S}}    \eps_{r,\mathcal{S}}^2 (  \ProjPF , \vCoeff_{*} )  \, .
\end{equation}
Because the errors from predicting a trajectory will likely be higher (snapshots capture distributions vs. trajectories of delta functions), we introduce a hyperparameter $\lambda\in(0,1)$. 

Finally, as we may be wishing to include many trajectories ($K_\mathcal{T} \gg 1$), having unassigned initial conditions for all trajectories will lead to an increase of $K_\mathcal{T} \LimBasis$ parameters. Therefore, we assign the initial condition to the first observation, i.e., $\vCoeff_{k,*} = \vect{c}_{0,k}$ and therefore $\eps_{0,\mathcal{T}} \equiv 0$, leading to loss function 
\begin{equation}
      \mathcal{L}_{\text{tot}} = \frac{\lambda}{ K_\mathcal{T} ( R_\mathcal{T}+1 )}  \sum_{k=1}^{K_\mathcal{T}} \sum_{r=1}^{R_\mathcal{T}}    \eps_{r,\mathcal{T}}^2 (  \ProjPF  ) +  \frac{1-\lambda}{R_\mathcal{S}+1}  \sum_{r=0}^{R_\mathcal{S}}    \eps_{r,\mathcal{S}}^2 (  \ProjPF , \vCoeff_{*} )  \, .
\end{equation}

\section{Experiments}\label{sec_workedexample}

\subsection{Simulated Data Generation}

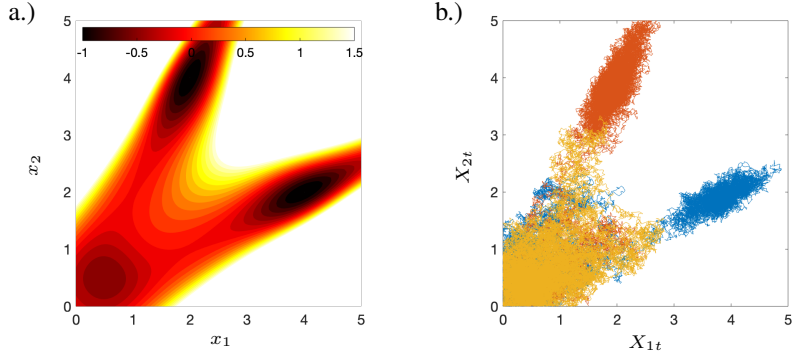
\begin{figure}
 \begin{center}
  \begin{overpic}[width=0.4\textwidth]{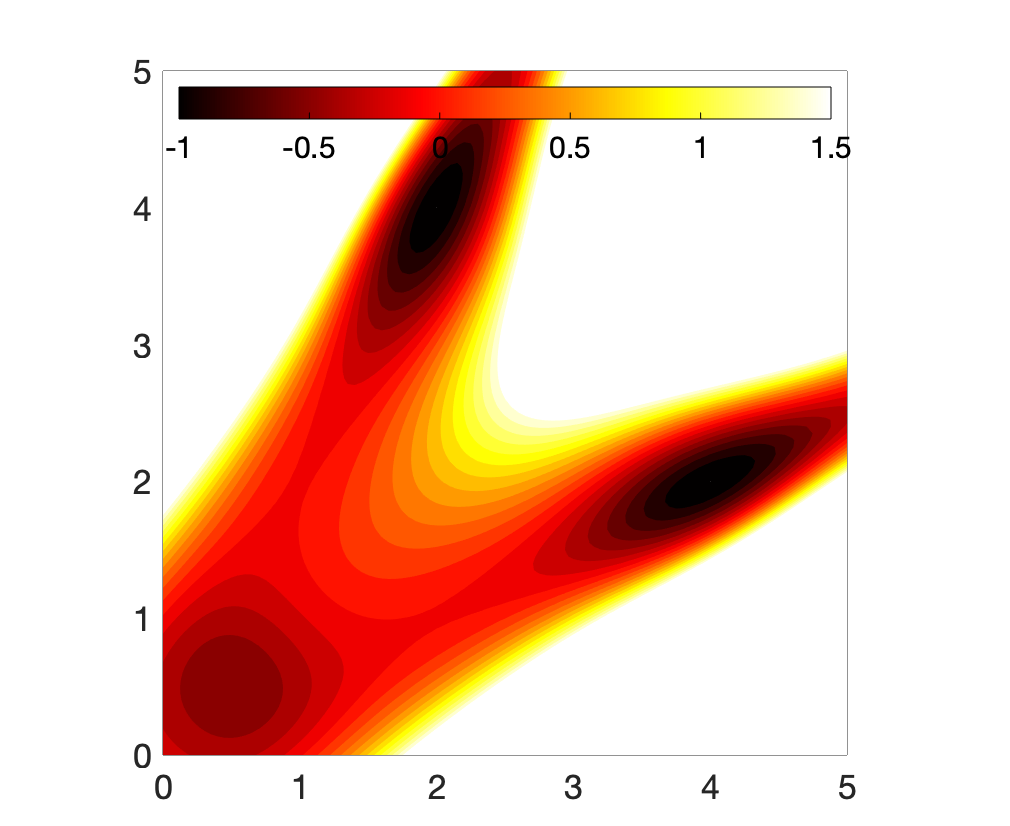}
   \put(0,75){a.)}
        \put(48,-2){\scriptsize $x_1$}
    \put(5,38){\scriptsize \rotatebox{90}{$x_2$}}
  \end{overpic}
    \begin{overpic}[width=0.4\textwidth]{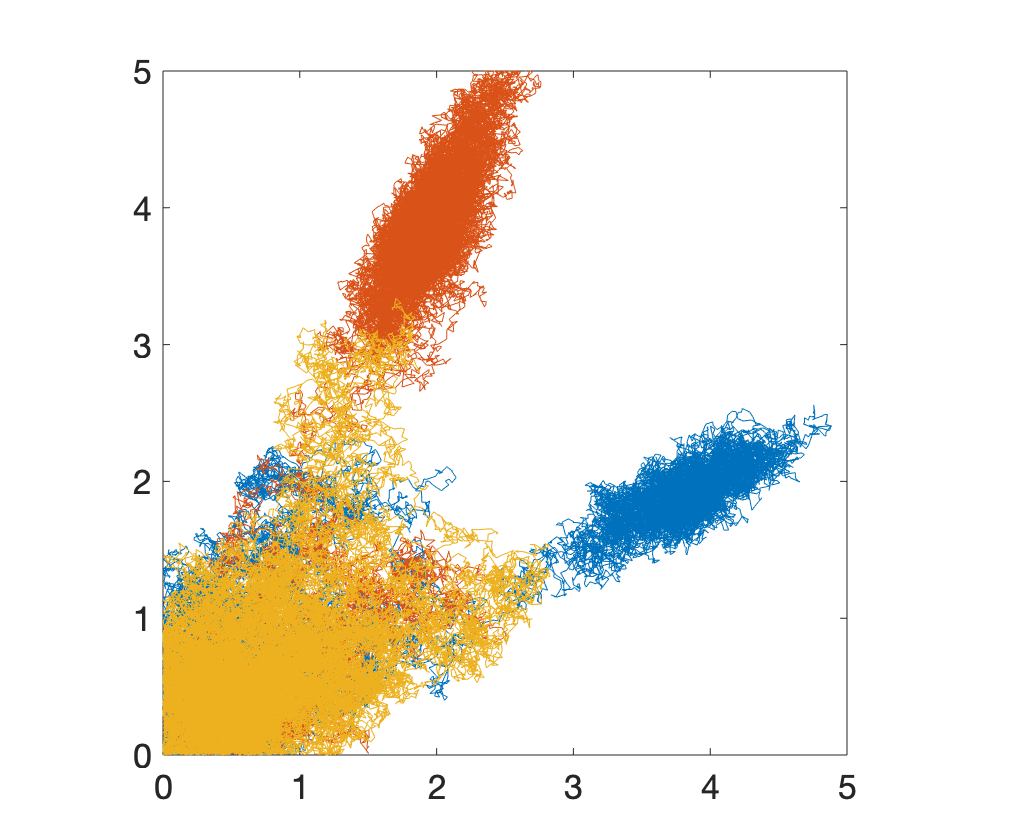}
   \put(0,75){b.)}
        \put(46,-3){\scriptsize $X_{1t}$}
    \put(5,36){\scriptsize \rotatebox{90}{$X_{2t}$}}
  \end{overpic}
    \end{center}
  \caption{
    \footnotesize{(a.) The potential well from equation \eqref{eq_pot_well} is illustrated. (b.) Three sample trajectories are shown from the stochastic dynamical system described by equation \eqref{eq_2d_SDE}--\eqref{eq_pot_well}.}}
  \label{fig_SDE_pot_well}
\end{figure}

For this worked example, we take the stochastic differential equation from \citet{taylor2020dynamic}. We consider point particles in a bistable potential well undergoing Brownian fluctuations. After initialisation around point $\Transpose{(1,1)}/2$, particles may switch between bistable paths along lines: $y = 2x$ or $y = x/2$ to finally settle in one of the two final state $\Transpose{(2,4)}$ or $\Transpose{(4,2)}$. The process is described by two-dimensional SDE $\{ \vect{X}_t = \Transpose{(X_{1t},X_{2t})} \in\Real^2_+ : t \geq 0 \}$
\begin{align}\label{eq_2d_SDE}
\dd \vect{X}_t = - \nabla V( \vect{X}_t ) \dd t + \sqrt{2D}\, \dd \vect{W}_t \, ,
\end{align}
where $\vect{W}_t$ is a two-dimensional Wiener process and the potential well is plotted in Figure \ref{fig_SDE_pot_well}(a) and is given by
\begin{align}\label{eq_pot_well}
V( \vect{x} ) &=   \left( \frac{1}{2}[\Vert \vect{x}\Vert^2 - x_1 x_2] - \frac{1}{10} \Vert \vect{x}\Vert^2  \right)^2 -\frac{1}{2} e^{- \frac{1}{2} \Vert \vect{x} - \Transpose{(\frac{1}{2},\frac{1}{2})} \Vert^2 } -   e^{- \Vert \vect{x} - \Transpose{(4,2)} \Vert^2 } -   e^{ - \Vert \vect{x} - \Transpose{(2,4)} \Vert^2 }
 \, .
\end{align}
As an initial condition at $\tp{0} =0$, the sample is placed with a multivariate normal distribution with mean $\vect{\mu} = \Transpose{(1,1)}/2$ and covariance matrix $\Sigma = \text{diag}(0.5,0.5)$. Three sample trajectories are visualised using an Euler--Maruyama with  time step $\de t = 2^{-9}$, see Figure \ref{fig_SDE_pot_well}(b); the diffusion constant is chosen to be $D = 1/4$. Along the lines $x=0$ and $y=0$, the system has reflecting boundary conditions. 

\subsubsection{Datasets Generated}\label{sec_datasets}
To limit the introduction of sampling errors, all datasets are subsets of a parent dataset whereby 1000 trajectories generated by the SDE described by equation \eqref{eq_2d_SDE}--\eqref{eq_pot_well} are observed at time points $t=0$, $1$, $2$, $3$, $5$, $8$, $13$, $21$, $34$, $55$ and $89$.
\paragraph{Trajectory Dataset}
For this dataset, 100 samples are randomly taken at the 11 timepoints listed above. As one can imagine, the early trajectories with small gaps between observations contain the most relevant information.
\paragraph{Full Snapshot Dataset}
Here, all 1000 samples are used. Of course, continuous trajectories are not used --- therefore for the samples associated to a fixed timepoint, index permutation does not change the results.
\paragraph{Two Snapshot Dataset}
A reduced snapshot dataset is created by taking the first and fourth snapshot. Most of the fast dynamics have taken place by this point in time.
\paragraph{Integrated Dataset}
A composite dataset is created by taking both the Two Snapshot Dataset and 10 of the 100 samples from the Trajectory Dataset.

\subsection{Basis Functions Selection} 
For Sparse DDD, we require compact local basis functions. For simplicity, we use radial piecewise linear basis functions
\begin{equation}
    \Basis_j (\vect{x}) = \frac{d(d+1)\Gamma (\frac{d}{2})}{2 \pi^{d/2} \zeta_j^d  } \left\{ \begin{array}{ccc}
        1-\frac{\norm{\vect{x} - \vect{x}_j}}{ \zeta_j } & \text{for} &  \norm{\vect{x} - \vect{x}_j} < \zeta_j \\
         0 &  & \text{otherwise}\, .
    \end{array} \right.
\end{equation}
We select the values of $\{\vect{x}_j\}_{j=1}^{\LimBasis}$ using $k$-means clustering with $k=30$ over the Full Snapshot Dataset. The values of $\zeta_j$ are chosen such that the support of each basis function covers the centers of the 2 nearest basis functions.
\begin{figure}[h]
\begin{mdframed}[style = tag]
\begin{center}
\subsection*{Mass Matrix Calculation}
\end{center}
The mass matrix is given by 
\begin{equation}\label{eq_M_def}
    M_{ij} = \InProd{\Basis_i, \Basis_j}_{\mu} = \int_{\MeaSp} \Basis_i(\State) \Basis_j (\State) \,  \dd \mu( \State ) \, .
\end{equation}
Typically in the Dynamic Mode Decomposition field, one would use a data driven measure $\mu_{D}( \State ) = \sum_{l = 1}^L \de (\State - \State_l )/L$ and then $M_{ij}$ is calculated as
\begin{equation}
    M_{ij} = \int_{\MeaSp} \Basis_i(\State) \Basis_j (\State) \,  \dd \mu_D ( \State ) = \frac{1}{L}\sum_{l = 1}^L \Basis_i(\State_l) \Basis_j (\State_l) \, .
\end{equation}
However, for a combination of snapshot and trajectory data one may have different numbers of data points $k_r$ at each time point $\tp{r}$. This may complicate the use of a data driven measure. Therefore, we integrate equation \eqref{eq_M_def} as a Riemann integral. As analytical expressions for many choices in basis functions are seldom known for equation \eqref{eq_M_def}, we use Monte Carlo integration  
\begin{equation}\label{eq_M_MC}
M_{ij} = \int_{\MeaSp} \Basis_i(\State) \Basis_j (\State) \,  \dd \State  \approx \frac{ |\Omega | }{L}\sum_{l = 1}^L \Basis_i(\State_l) \Basis_j (\State_l) \, ,
\end{equation}
whereby $\State_l$ is sampled uniformly at random from $\Omega \subset \mathds{R}^d$. We choose $\Omega$ as a $d$-dimensional hypercube to enclose the support of one of the basis functions. Superior convergence of equation \eqref{eq_M_MC} is achieved via quasi-Monte Carlo integration with Halton node placing \cite{morokoff1995quasi}.
\end{mdframed}
\vspace{-1em}
\end{figure}

\subsection{Comparison between DDD and Sparse DDD}
To prevent numerical instability for large times, we re-scale the experimental time course $(\tp{1},\tp{R})$ to the unit interval. We then run DDD (Section \ref{sec_DDD}) and Sparse DDD (Section \ref{sec_SDDD}) on the 4 datasets created in Section \ref{sec_datasets} using sequential quadratic programming (SQP).

For the $\LimBasis = 30$ basis functions, DDD has $\LimBasis^2 = 900$ parameters or $\LimBasis(\LimBasis-1) = 870$ degrees of freedom (columns of $\PFmat$ summing to zero). However, after fixing the corresponding zero-valued indices of $M$ to zero in $Q$, Sparse DDD has 220 parameters, or 190 degrees of freedom. As one would expect, Sparse DDD is computationally much faster than DDD.

\begin{table}
  \caption{Mean relative error for Sparse DDD and DDD applied to the 4 datasets from Section \ref{sec_datasets} and tested on the Full Snapshot Dataset.}
  \label{tab_error_comp}
  \centering
  \begin{tabular}{lll}
    \toprule
    & \multicolumn{2}{c}{Mean relative error} \\
    \cmidrule(r){2-3}
    Dataset     & Sparse DDD     & DDD \\
    \midrule
    Trajectory    & 0.760 & 0.811  \\
    Full Snapshot & 0.079 & 0.053  \\
    Two Snapshot  & 0.360 & 0.303  \\
    Integrated     & 0.291 & 0.239  \\
    \bottomrule
  \end{tabular}
\end{table}

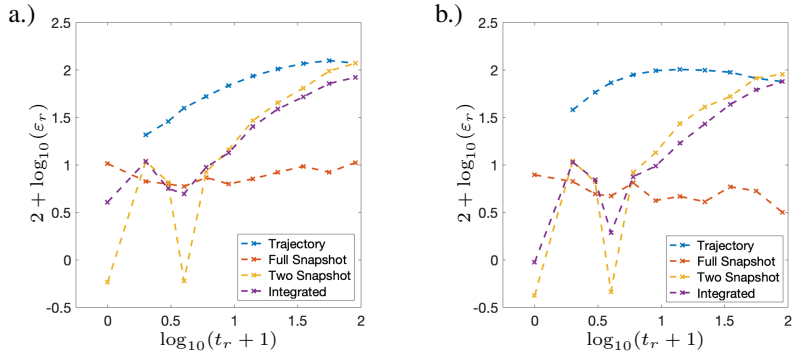
\begin{figure}[h]
  \begin{center}
  \begin{overpic}[width=0.40\textwidth]{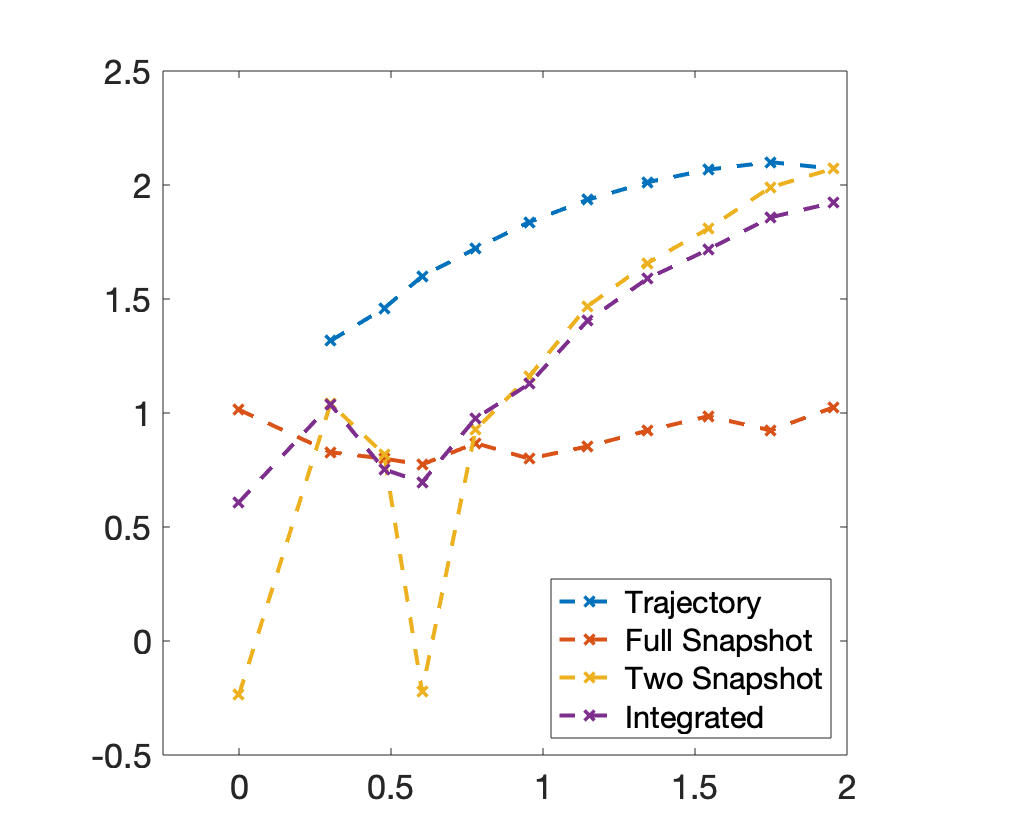}
   \put(0,75){a.)}
    \put(36,-2){\scriptsize $\log_{10}(\tp{r} + 1)$}
    \put(5,27){\scriptsize \rotatebox{90}{$2 + \log_{10}(\eps_r)$}}
  \end{overpic}
    \begin{overpic}[width=0.40\textwidth]{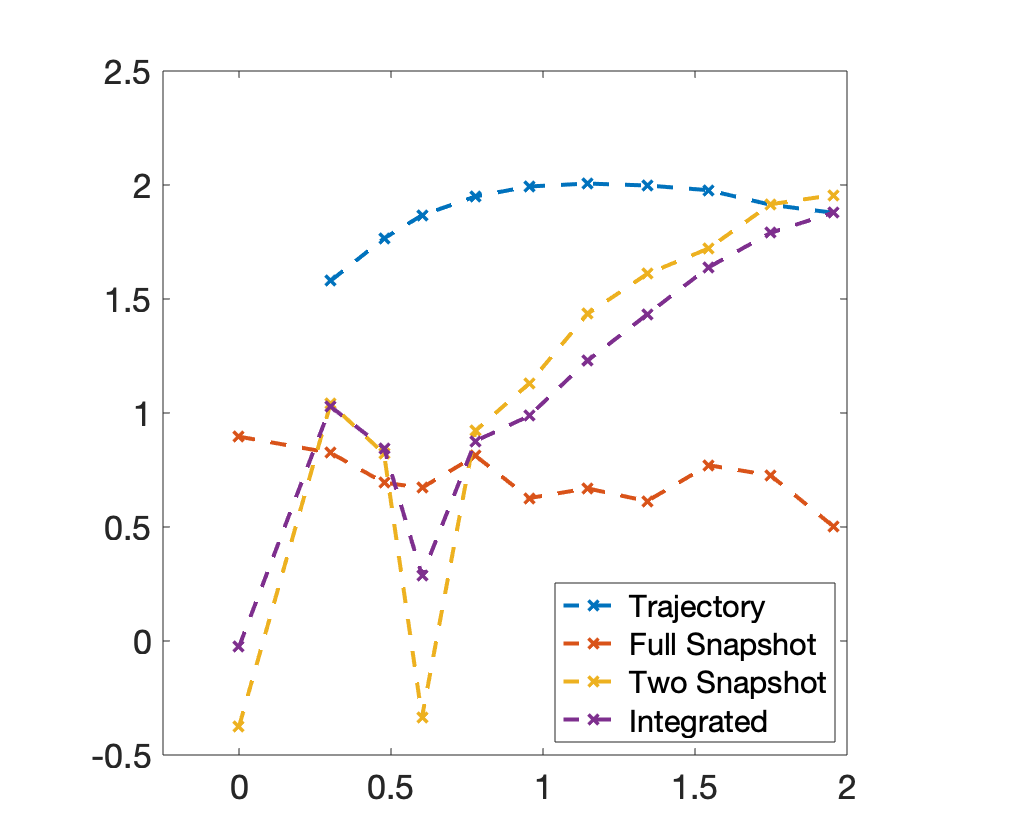}
   \put(0,75){b.)}
    \put(36,-2){\scriptsize $\log_{10}(\tp{r} + 1)$}
    \put(5,27){\scriptsize \rotatebox{90}{$2 + \log_{10}(\eps_r)$}}
  \end{overpic}
    \end{center}
  \caption{
    \footnotesize{Relative prediction error for the Full Snapshot dataseat each time point for: (a.) Sparse DDD, and (b.) DDD.}}
  \label{fig_error_comp}
\end{figure}

In Table \ref{tab_error_comp}, we show the mean relative error for the test Full Snapshot Dataset --- but trained separately on the 4 individual datasets. We find that, apart from the Trajectory Dataset, DDD always achieves a marginally lower error. We also plot the log percentage relative predictive error at each time point in Figure \ref{fig_error_comp}.

Examining Figure \ref{fig_error_comp}, we find that:
\begin{itemize}
\item The Trajectory Dataset has consistently high error, but this is for two reasons: (1) due to the large time gaps between observations,  100 trajectories does not contain a huge amount of information to infer a Markov transition rate matrix; and (2) as explained in Section \ref{sec_integration}, initial conditions for time series are not left as a free parameter. For evaluation of the test set, we fix the initial condition as $\vCoeff_{*} = \vect{c}_{0}$. 
\item The Full Snapshot Dataset achieves consistently low fitting error. Note that for this problem, the training and test set are identical.
\item  The Two Snapshot Dataset achieves very low error around the specified two time points, but error quickly increases for later time points. 
\item  By incorporating 10 trajectories, the Integrated Dataset achieves better error  at later time points than the Two Snapshot Dataset. For the free hyperparameter from Section \ref{sec_integration}, we choose $\lambda = 0.01$.  
\end{itemize}
Taking these observations together, we show that information from trajectory and snapshot time series data can be combined to provide better fits than either datasets individually.

\section{Discussion and Conclusion}\label{sec_discussion}

We have presented an approach for synergistic integration of trajectory and snapshot time series datasets. We believe that data integration of this form will become increasingly relevant as we wish to: (1) integrate different patient datasets (static vs. longitudinal cohorts) and (2) integrate various modalities of cell profiling, such as live cell imaging with accompanied multi-omics. A key future challenge will be methods designed with incomplete observations using different measurements, for example, single cell RNA-seq ($\sim$20,000 mRNA transcripts) and flow/mass cytometry ($\sim$5-50 surface proteins).

The numerical approximation of operators is an emerging field and therefore approaches are still being optimised and refined. A hurdle practitioners often face is the selection of basis functions that well characterise the time series data; many options are available, including radial basis functions \citep{fornberg2015primer} and global basis functions \citep{klus2016On}. Greater numbers of basis functions alleviate the problem of data better data representation at the cost of more parameters to be inferred. Sparse methods have therefore been proposed for DMD in a discrete time setting \cite{dicle2016robust, kaneko2019convolutional, Heas2017Optimal}, but Sparse DDD appears to be a first attempt in a continuous-time setting.

\begin{ack}

We gratefully acknowledge the help of Tomislav Plesa, Asbjorn Riseth and Manfred Claassen. 

This work has been wholly supported by Relation Therapeutics Limited. 

\end{ack}

\medskip

\small

\bibliographystyle{natbib}
\bibliography{References_22Nov2017.bib,References_BioRefFromManfred.bib,DDD.bib}

\appendix

\end{document}